\documentclass[default]{sn-jnl}

\usepackage{lmodern}

\jyear{2021}%

\theoremstyle{thmstyleone}%
\usepackage[table]{xcolor}
\usepackage{pifont}
\newcommand{\cmark}{\ding{51}}%
\usepackage{booktabs}   
\usepackage{tabularx}    
\newcolumntype{Y}{>{\centering\arraybackslash}X}  

\theoremstyle{thmstylethree}%

\begin{document}

\title[Personalized Federated Learning via Dual-Prompt Optimization and Cross Fusion]{Personalized Federated Learning via Dual-Prompt Optimization and Cross Fusion}

\author[1,2]{\fnm{Yuguang} \sur{Zhang}}

\author[1]{\fnm{Kuangpu} \sur{Guo}}

\author[3]{\fnm{Zhihe} \sur{Lu}}

\author[4]{\fnm{Yunbo} \sur{Wang}}

\author[1,2]{\fnm{Jian} \sur{Liang}}

\affil[1]{\orgname{Institute of Automation, Chinese Academy of Sciences}, \country{China}}

\affil[2]{\orgname{University of Chinese Academy of Sciences}, \country{China}}


\affil[3]{\orgname{Hamad Bin Khalifa University},  \country{Qatar}}

\affil[4]{\orgname{Central South University},  \country{China}}

\abstract{
Federated learning (FL) enables collaborative model training across decentralized clients without sharing local data, but is challenged by heterogeneity in data, computation, and communication. 
Pretrained vision-language models (VLMs), with their strong generalization and lightweight tuning via prompts, offer a promising solution. 
However, existing federated prompt-learning methods rely only on text prompts and overlook joint label–domain distribution shifts.
In this paper, we propose a personalized FL framework based on dual-prompt learning and cross fusion, termed \textbf{pFedDC}. Specifically, each client maintains both global and local prompts across vision and language modalities: global prompts capture common knowledge shared across the federation, while local prompts encode client-specific semantics and domain characteristics.
Meanwhile, a cross-fusion module is designed to adaptively integrate prompts from different levels, enabling the model to generate personalized representations aligned with each client’s unique data distribution. 
Extensive experiments across nine datasets with various types of heterogeneity show that pFedDC consistently outperforms state-of-the-art methods. 
}


\maketitle

\section{Introduction}\label{sec1}

With the increasing emphasis on data privacy, federated learning (FL)~\cite{mcmahan2017communication} has emerged as a promising paradigm that enables decentralized model training without requiring participants to share their local data.
Despite its advantages, FL faces significant challenges due to data heterogeneity—such as label and domain shifts—which hinder model convergence and degrade performance~\cite{cui2024harmonizing, li2024global, yu2023federated, guo2024not, luo2021no, shi2022towards, guo2024dynamic}.
Furthermore, conventional FL approaches often incur substantial computational and communication overhead owing to frequent parameter updates exchanged with the central server~\cite{li2024global}.
These limitations have generally confined FL applications to lightweight model architectures, resulting in suboptimal performance and training instability~\cite{yang2023efficient}.

Recently, pre-trained vision-language models, such as CLIP~\cite{radford2021learning} and ALIGN~\cite{jia2021scaling}, have demonstrated strong generalization across downstream tasks, making them well-suited for integration with FL.
This synergy enables privacy-preserving, communication-efficient training while benefiting from the representational strength of foundation models~\cite{cui2024harmonizing, li2024global}.
However, full model fine-tuning in FL settings introduces prohibitive communication costs and risks of overfitting, especially on the condition of client data is limited.
Prompt learning~\cite{liu2023pre} adapts large pre-trained models via a small set of tunable parameters and offers a lightweight alternative.
Consequently, recent studies have explored incorporating prompt learning into FL frameworks to enhance efficiency and performance~\cite{zhao2022reduce, guo2023promptfl, li2024global, cui2024harmonizing}.
For example, PromptFL~\cite{guo2023promptfl} directly applies FedAvg~\cite{mcmahan2017communication} to aggregate prompt parameters from clients.

In real-world scenarios, client data often exhibits domain shift~\cite{li2021fedbn} and label shift~\cite{li2021model}, posing significant challenges to federated learning.
Directly applying FedAvg~\cite{mcmahan2017communication} to prompt tuning frequently results in divergence from local data distributions and suboptimal performance.
To this end, personalized federated prompt learning approaches are necessary to effectively manage data heterogeneity.
pFedPrompt~\cite{guo2023pfedprompt} introduces personalization by maintaining client-specific attention modules for generating localized spatial visual features, while leveraging shared text prompts to capture global consensus.
Similarly, FedOTP~\cite{li2024global} employs unbalanced optimal transport to align local visual features with both global and local text prompts, enabling the model to represent diverse category characteristics across clients.
However, existing federated prompt learning methods predominantly focus on text prompt adaptation and lack comprehensive strategies to systematically address complex and multifaceted data heterogeneity.

To address this issue, we propose \textit{Personalized FL framework Based on Dual-
prompt Learning and Cross Fusion} (pFedDC). 
In heterogeneous data environments—particularly under domain shift—visual features corresponding to the same category may vary significantly across clients, impeding global model learning~\cite{bai2024diprompt}.
To mitigate the effects of label and domain shifts, pFedDC simultaneously learns shared global text and vision prompts, along with personalized local text and vision prompts for each client during the local training phase.
The vision prompts enhance robustness to visual misalignment and support domain adaptation, while the text prompts capture category-level semantics~\cite{wei2023dual}.
After local training, local prompts are retained on each client, whereas global prompts are transmitted to the server and aggregated using FedAvg~\cite{mcmahan2017communication}.
This strategy enables clients to acquire shared knowledge via global prompts while maintaining local adaptability through personalized prompts.

To further integrate global and local information in a personalized and effective manner, pFedDC employs cross-attention modules for both text and vision prompts.
The text cross-attention module takes global and local text prompts as input and generates fused text representations, which are then fed into the text encoder to extract discriminative category-specific features.
Similarly, the vision cross-attention module operates on vision prompts to produce refined visual representations that reflect both global context and local domain characteristics.
These cross-attention modules are updated jointly with the prompts and are maintained locally to preserve client-specific adaptations.
Extensive experiments on federated prompt learning benchmarks involving various types of data heterogeneity—including both label and domain shifts—demonstrate that pFedDC achieves superior performance compared to existing methods. 
The results validate the effectiveness of our dual-prompt design and cross-attention mechanism, providing a scalable and privacy-preserving solution for real-world FL deployments.
Our main contributions are summarized as follows:

\begin{itemize}
\item We are the first to investigate the mechanism of dual-mode prompt tuning in federated learning under complex data heterogeneity. Specifically, we train text prompts to capture class information and vision prompts to learn domain-specific information simultaneously.
\item We propose a federated learning framework termed pFedDC, which leverages cross-attention modules to enhance the cooperation between global and local prompts across text and vision modalities. By fusing the local and global prompts, pFedDC learns personalized prompts that are better suited to the local data distribution.
\item We carried out comprehensive and thorough experiments on several widely recognized datasets, which encompass different forms of data heterogeneity, such as domain shift and label shift. The significant improvement in results demonstrates the superiority of our pFedDC.
\end{itemize}

\section{Related Work}\label{sec2}

\subsection{Personalized Federated Learning}
Personalized federated learning (PFL) is a highly regarded research field because of its potential to address statistical and systemic heterogeneity across clients.
Various approaches~\cite{guo2024addressing, xu2023personalized, ye2023feddisco, arivazhagan2019federated, collins2021exploiting, oh2021fedbabu, li2023fedtp, li2021fedbn, sun2021partialfed, li2021ditto, zhang2023fedcp, luo2024mixture, t2020personalized} have been proposed in prior works to achieve PFL.
FedPAC~\cite{xu2023personalized} and FedDisco~\cite{ye2023feddisco} introduce novel weighted aggregation techniques to promote intensive collaboration among similar clients.
FedPer~\cite{arivazhagan2019federated}, FedRep~\cite{collins2021exploiting}, and FedBABU~\cite{oh2021fedbabu} learn personalized classifier heads locally while sharing the base layers, and FedTP~\cite{li2023fedtp} learns personalized self-attention layers for each client.
FedBN~\cite{li2021fedbn} employs local batch normalization to mitigate the feature shift before model averaging and PartialFed~\cite{sun2021partialfed} extends this strategy by selecting personalized parameters according to distinct feature traits of different clients.
Moreover, Ditto~\cite{li2021ditto} further incorporates a regularization term to encourage local models to retain information from the global model.
FedCP~\cite{zhang2023fedcp} takes a data-centric perspective by decoupling features into global information features and personalized information features, which are processed by a global head and a personalized head for different tasks.
The latest pFedMoAP~\cite{luo2024mixture} personalizes the prompt learning process through the lens of Mixture of Experts.
Similarly, pFedMe~\cite{t2020personalized} employs Moreau envelopes as a regularization technique, enabling each client
to learn an additional personalized model.
The methods mentioned above primarily target label shift or domain shift data heterogeneity.
However, they may not perform well when label shift and domain shift exist at the same time.
Our pFedDC explores the cooperation between global and local prompts for both text and vision modalities to effectively address both label shift and feature shift.

\subsection{Federated Prompt Learning}
In recent years, prompt learning with vision–language models (VLMs) has been incorporated into the federated learning paradigm to reduce per-client computational overhead while simultaneously addressing two core challenges: robustness across heterogeneous domains and the non-IID nature of decentralized data~\cite{qiu2023text,li2024position,halbe2023hepco,su2022cross,yang2023efficient}. 
Early adaptations—such as FedPrompt~\cite{zhao2022reduce} and PromptFL~\cite{guo2023promptfl}—extend the standard fine-tuning procedure of CLIP~\cite{radford2021learning} to the federated setting, demonstrating that prompt-based adaptation can achieve performance on par with or superior to full-model updates.
Building on these foundations, subsequent works have advanced federated prompt learning in diverse application contexts.
FedPR~\cite{feng2023learning} constrains visual-prompt updates to lie within the null space of the global prompt for accelerated MRI reconstruction.
FedAPT~\cite{su2022cross} introduces an adaptive prompt-tuning algorithm for cross-domain image classification, whereas FedCLIP~\cite{lu2023fedclip} employs an attention-based adapter to better leverage pre-trained model features in each client.
To mitigate statistical heterogeneity, pFedPrompt~\cite{guo2023pfedprompt} equips each client with a non-parametric personalized attention module that synthesizes spatially adaptive visual prompts, and pFedPG~\cite{yang2023efficient} delegates prompt generation to a server-side client-specific generator for tailored local adaptation.
More recently, FedOTP~\cite{li2024global} applies unbalanced optimal transport to align local visual embeddings with both global and local textual prompt embeddings, thereby capturing fine-grained, per-client category characteristics.
Despite these advances, nearly all existing federated prompt-learning frameworks focus exclusively on tuning textual prompts, with minimal exploration of vision-prompt adaptation. 
This limitation hinders their ability to accommodate substantial label shifts or pronounced feature distribution drifts across clients, an important direction for future research.

\subsection{Attention Module}
It is well known that attention plays a crucial role in human perception. Recently, numerous studies~\cite{wang2017residual,hu2018squeeze,wang2018non,woo2018cbam,ramachandran2019stand,zhao2020exploring} have integrated attention mechanisms into convolutional neural networks (CNNs) to enhance performance on large-scale classification tasks.
RAN~\cite{wang2017residual} proposes a Residual Attention Network, which uses an encoder-decoder style attention module to iteratively refine feature representations. 
The Squeeze-and-Excitation (SE) block~\cite{hu2018squeeze} uses a lightweight gating mechanism to model inter-channel dependencies and boost representational power.
CBAM~\cite{woo2018cbam} exploits both spatial and channel-wise attention based on an efficient architecture and empirically verifies that exploiting both is superior to using only the channel-wise attention.
There has also been a lot of interest in combining CNNs with different forms of self-attention.
The non-local neural network~\cite{wang2018non} introduces a global attention operation by computing pairwise correlations across all spatial locations for dense contextual aggregation.
Local self-attention variants such as Stand-Alone Self-Attention (SASA)~\cite{ramachandran2019stand} and the Self-Attention Network (SAN)~\cite{zhao2020exploring} replace standard convolutions with attention layers confined to local neighborhoods, capturing fine-grained spatial interactions.
In this paper, we propose to incorporate cross-attention modules that fuse global consensus and local personalization across both vision and language modalities, thereby achieving an optimal balance between shared context and client-specific adaptation.  

\begin{figure}[t]
    \centering
    \includegraphics[width=\textwidth]{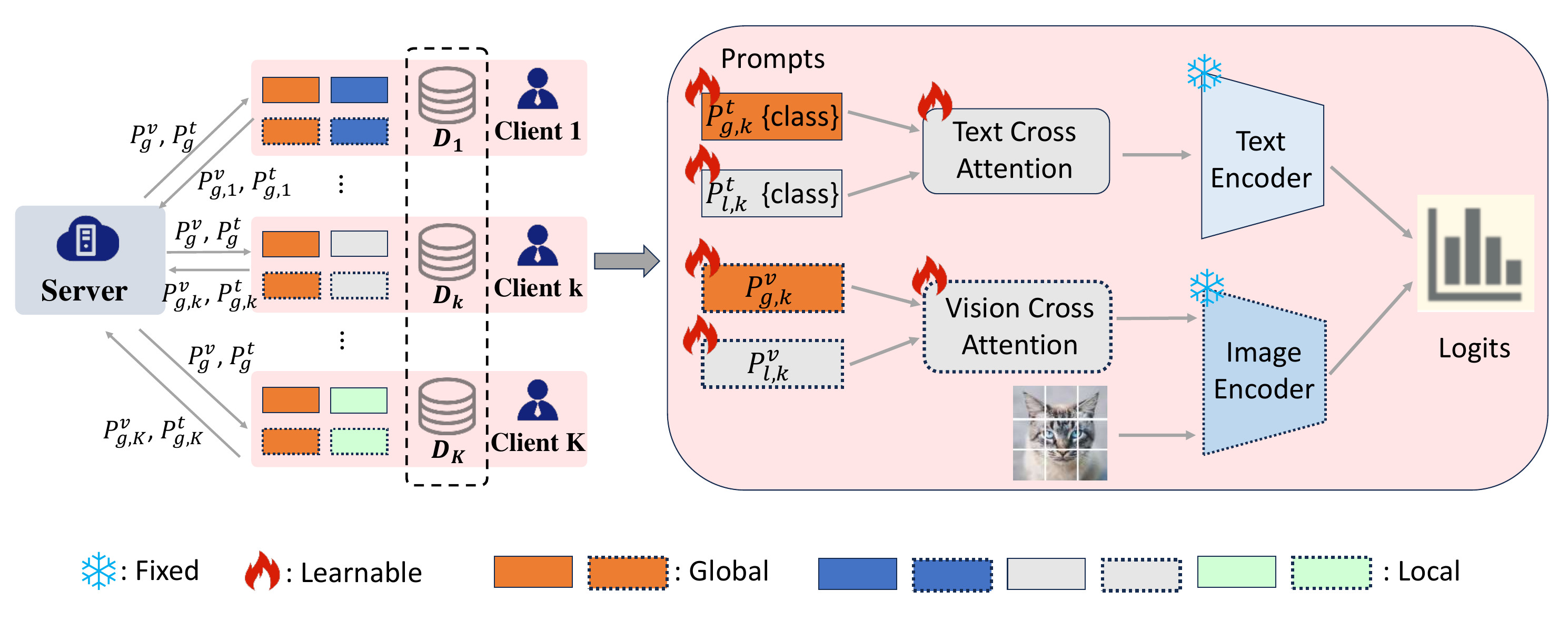}
    \caption{
    \textbf{The overview of our pFedDC.} On the left, clients transmit global text and vision prompts to the server for aggregation while retaining local text and vision prompts locally. The right shows the workflow of global and local prompts cooperation with the cross-attention mechanism.}
    \label{fig:method-client}
\end{figure}

\section{Methods}\label{sec3}
In this section, we present a detailed overview of the research problem and our proposed method.
First, in Sec.~\ref{Preliminaries}, we review foundational concepts related to CLIP~\cite{radford2021learning} with prompt tuning and a standard federated learning framework.
We then introduce our proposed method, pFedDC, which is built upon dual prompt learning and cross-attention for global and local prompts, detailed in Sec. 3.2 and Sec. 3.3 respectively.

\subsection{Preliminaries}
\label{Preliminaries}
In this paper, we adopt CLIP~\cite{radford2021learning} as the foundational model.
CLIP employs a dual-branch architecture comprising an image encoder, $f^v(\cdot)$, and a text encoder, $f^t(\cdot)$, with each encoder designed to process data from its respective modality.
For zero-shot predictions in downstream tasks, CLIP~\cite{radford2021learning} utilizes a human-designed prompt (e.g., “a photo of a [CLASS]”) for each class.
Taking a C-way classification task as an example, the textual embeddings of all classes $\{f^t_c\}_{c=1}^C$ and the embedding of the test image $f^v(x)$ are generated by the text and image encoders, respectively.
The probability that a given image $x$ belongs to the $c$-th category is determined by applying the softmax operation as follows:
\begin{equation}
p_c(x) = \frac{\exp(\text{sim}(f^v(x), f^t_c)) / \tau)}{\sum_{j=1}^{C} \exp(\text{sim}(f^v(x), f^t_j)) / \tau)},
\end{equation}
where $\tau$ is a temperature parameter and C is the total number of categories.
To improve performance in downstream tasks, prompt tuning has been widely adopted as a parameter-efficient fine-tuning approach~\cite{zhou2022learning}.
This involves introducing additional learnable textual tokens, $P^t$, and visual tokens, $P^v$ (referred to as text/vision prompts) into the corresponding encoders, thereby optimizing the original CLIP model for specific downstream tasks~\cite{ jia2022visual, xing2023dual}.

\begin{figure}
    \small
    \setlength\tabcolsep{1mm}
    \renewcommand\arraystretch{0.1}
    \begin{tabular}{cc}
        \includegraphics[width=0.475\linewidth]{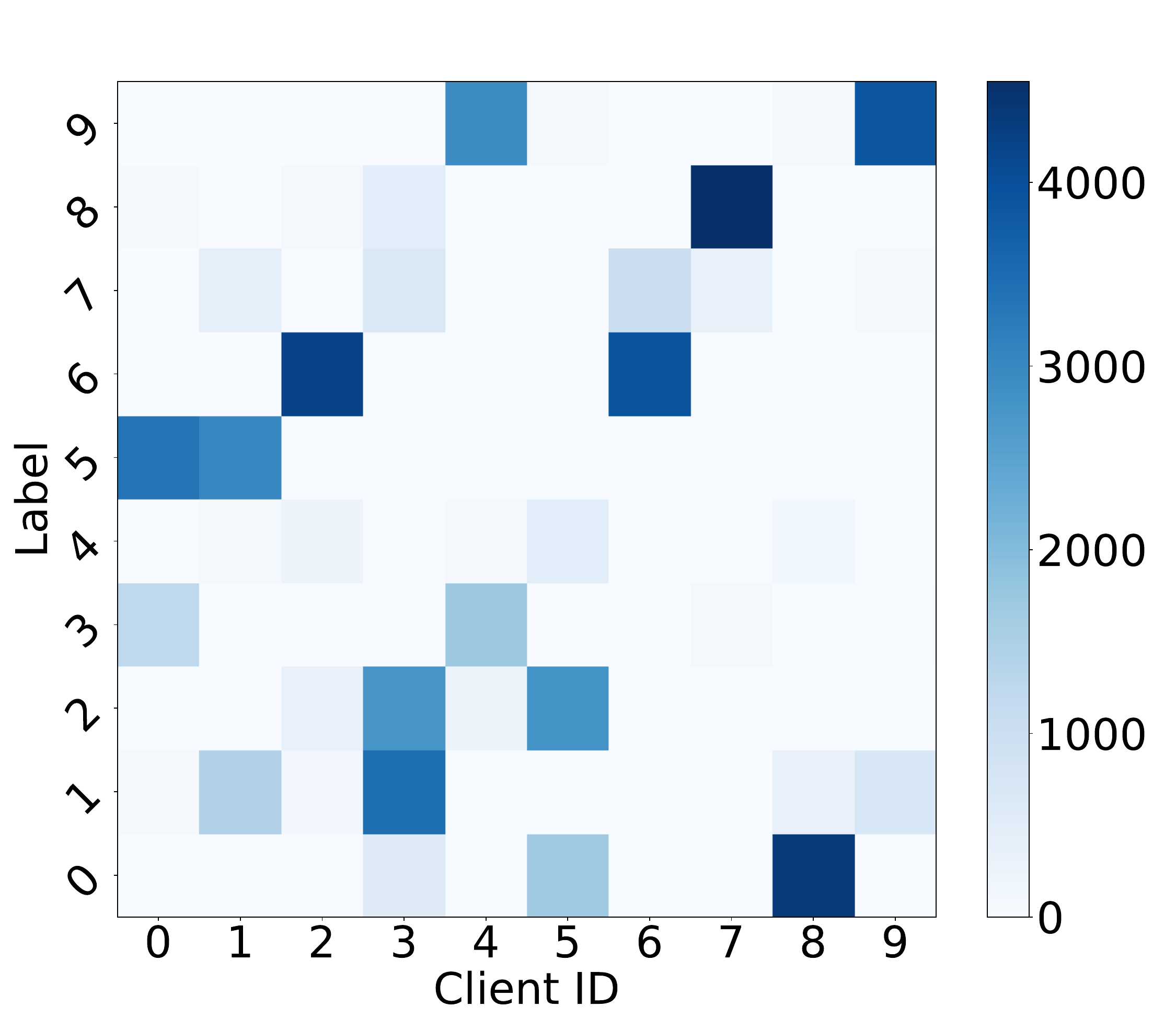} & 
        \includegraphics[width=0.5 \linewidth, clip]{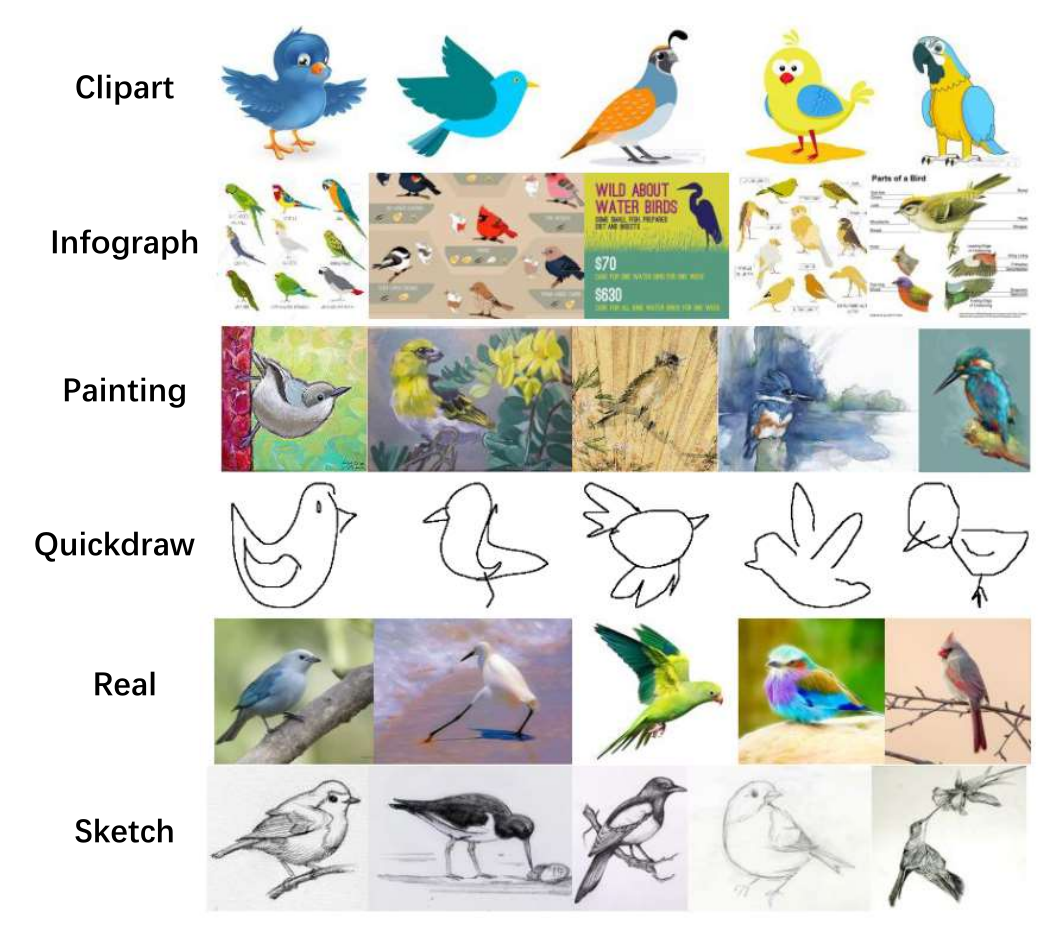} 
    \\[1.0mm]
        (a) Label shift. & (b) Domain shift.
    \end{tabular}
    \caption{Illustration of data heterogeneities. (a) Label distribution shift among 10 clients with 10 classes.  Darker colors indicate that a client holds a larger number of training samples for the corresponding class. (b) Domain shift of Domainnet.}
    \label{fig: shift}
\end{figure}

In federated learning, we consider a scenario with $K$ clients, where $\mathcal{D}_k$ represents the local training data of client $k$ and contains $n_k$ samples.
The combined data $\mathcal{D}=\bigcup_{k=1}^{K} \mathcal{D}_k$ comprises the local data from all clients and the number of samples in $\mathcal{D}$ is $n = \sum_{k=1}^{K} n_k$. 
These data distributions might differ across clients, encompassing label and domain shift as shown in Fig.~\ref{fig: shift}.

\subsection{Personalized Federated Dual Prompts Learning}
\label{pFedDC1}
Data heterogeneities, including label shift and domain shift, present significant challenges in federated learning.
Previous federated prompt learning methods have primarily focused on text prompt tuning, overlooking vision prompt tuning, which is insufficient for addressing complex data heterogeneity.
To overcome this, we propose personalized federated learning with dual prompts and cross-attention, where each client learns both vision and text prompts, and cross-attention is used to fuse global and local prompts.

The process of feeding the text prompt into the text encoder can be viewed as synthesizing a classifier, while the image encoder extracts the visual features~\cite{xing2023dual}.
The prompting mechanism aims to query the relevant information, which is beneficial to the downstream tasks, from the pre-trained model.
However, previous methods adjust only the classifier while keeping the highlighted visual features unchanged.
Therefore, we believe that using prompt tuning only for the text encoder while keeping the image encoder fixed in federated learning with data heterogeneity is suboptimal, particularly in environments with complex domain shifts.

Therefore, we propose that clients simultaneously learn both the text prompt and the vision prompt.
To obtain a model that is better adapted to the client's local data distribution, personalization is also crucial.
During local training, each client learns both shared global text and vision prompts, as well as local text and vision prompts.
This approach enables clients to extract more individualized insights while maintaining a level of consensus among them.
Specifically, for client $k$, the prompt $P_k$ comprises a global text prompt $P_{g,k}^t$, a local text prompt $P_{l,k}^t$, a global vision prompt $P_{g,k}^v$, and a local vision prompt $P_{l,k}^v$, denoted as $P_k = [P_{g,k}^t, P_{l,k}^t, P_{g,k}^v, P_{l,k}^v] $.
During each communication round $r$, client $k$ initializes the prompts as $P_k^r = [P_{g}^{t, r-1}, P_{l,k}^{t, r-1}, P_{g}^{v, r-1}, P_{l,k}^{v, r-1}] $.
Then the global and local prompts are jointly updated through gradient descent $P_k^r = P_k^r - \eta \nabla \mathcal L_{D_i} (P_k^r)$ for E iterations locally.
After local training, only the updated global prompts $[P_{g,k}^{t, r}, P_{g,k}^{v,r}]$ will be transmitted to the server for aggregation to learn global consensus among clients, while the local prompts $[P_{l,k}^{t, r}, P_{l,k}^{v, r}]$ are retained locally to capture client-specific category characteristics.

In the server, we follow FedAvg~\cite{mcmahan2017communication} to aggregate global prompts as follows:
\begin{align}
P_{g}^{t,r+1} = \sum\limits_{k=1}^K \frac{n_k}{n} \cdot P_{g,k}^{t,r} \label{eq:aggregate_text}, \\
P_{g}^{v,r+1} = \sum\limits_{k=1}^K \frac{n_k}{n} \cdot P_{g,k}^{v,r} \label{eq:aggregate_vision}.
\end{align}
where $n_k$ is the number of samples in client $k$ and $n$ is the number of total samples in all clients.
We then transmit the aggregated global text and visual prompts to each client to initialize the client’s local prompts.

\begin{algorithm}[h]
    \renewcommand{\algorithmicrequire}{\textbf{Input:}}
    \renewcommand{\algorithmicensure}{\textbf{Output:}}
    \caption{\textbf{pFedDC}}
    \label{alg:pFedDC}
    \begin{algorithmic}[1]
        \Require  communication rounds $T$, client number $K$, local datasets $\{D_k\}_{i=k}^K$, sample counts $\{n_k\}_{k=1}^K$, client participating rate $R$, learning rate $\eta$, Softmax temperature $\tau$
        \Ensure  global text and vision prompts $P^{t,T}_g$ and $P^{v,T}_g$, personalized local text and vision prompts $\{P^{t,T}_{l,k}\}^K_{k=1}$ and $\{P^{v,T}_{l,k}\}^K_{k=1} $, personalized local text and vision cross-attention  $\{A^t_k\}^K_{k=1}$ and $\{A^v_k\}^K_{k=1}$
        \State   Initialize $P_k^0 \gets [P_{g}^{t, 0}, P_{l,k}^{t, 0}, P_{g}^{v, 0}, P_{l,k}^{v, 0}]$, $A^0_k \gets [A^{t, 0}_k, A^{v, 0}_k]$ for all $k=1,\dots,K$
        \State $m \gets \max( \lfloor R \cdot K \rfloor, 1)$
        \For{communication round $r=1,2,\cdots,T$}
            \State $M \gets $ Randomly select a subset containing $m$ clients.
            \State \textcolor{gray}{\# \textit{Local update}}
            \For{each client $k \in M$}
                \State Initialize local states:
                \State \quad $P_k^r \gets P_k^{r-1}$  \textcolor{gray}{\# \textit{$P_k^r$ denotes the set of prompts in client k }}
                \State \quad $A_k^r \gets A_k^{r-1}$ \textcolor{gray}{\# \textit{$A_k^r$ denotes the set of cross-attention modules in}}
                \State \quad \quad \quad \quad \quad \quad \quad \textcolor{gray}{\textit{ client k}}

                \For{\textit{each batch} $\mathcal{B}_{i} = \{\textit{x},{\textit{y}}\} \in {D}_k$} \vspace{0.05\baselineskip}  
                    \State Obtain fused local text/vision 
                    \State prompts with Eq. (\ref{eq:fuse_text})/Eq. (\ref{eq:fuse_vision}).           
                    \State $P^{r}_k \leftarrow P^{r}_k - \eta \nabla {\mathcal{L}({P^{r}_k;\mathcal{B}_i)}}$ 
                    \vspace{0.3\baselineskip}
                    \State $A^{r}_k \leftarrow A^{r}_k - \eta \nabla {\mathcal{L}({A^{r}_k;\mathcal{B}_i)}}$ 
                    \State \textcolor{gray}{\# $\mathcal{L}$ \textit{is cross-entropy loss}}
                \EndFor
            \EndFor
            \State \textcolor{gray}{\# \textit{Global aggregation}}
            \State Obtain aggregated text and vision prompts \State $P_{g}^{t,r}$ and $P_{g}^{v,r}$ with Eq. (\ref{eq:aggregate_text}) and Eq. (\ref{eq:aggregate_vision}).
        \EndFor
    \end{algorithmic}
\end{algorithm}

\subsection{Fusing Global and Local Prompts with Cross-attention}
\label{pFedDC2}
To further balance global consensus and local personalization, we introduce cross-attention modules for each client to fuse global and local prompts across both vision and text modalities.
Specifically, during local training, client $k$ is equipped with a text cross-attention module $A^t_k$ and a vision cross-attention module $A^v_k$, which are combined as $A_k = [A^t_k, A^v_k]$.
In each cross-attention module, the keys (K), queries (Q), and values (V) are computed through a linear transformation.
The fused personalized text and vision prompts in client $k$ are computed as follows: 
\begin{align}
P_{k}^{t,r} & = A^t_k(P_{g,k}^{t,r}, P_{l,k}^{t,r}) \label{eq:fuse_text}, \\
P_{k}^{v,r} & = A^v_k(P_{g,k}^{v,r}, P_{l,k}^{v,r})\label{eq:fuse_vision}.
\end{align} 
The fused prompts are fed into the frozen text and vision encoder to obtain the text and vision embeddings.
To adapt the fused prompts to the client's label distribution and domain characteristics, the text cross-attention module $A^t_k$ and vision cross-attention module $A^v_k$ are updated with the prompts and retained locally.
Moreover, keeping the cross-attention module local and personalized helps reduce communication overhead.
As a result, the objective function of our proposed pFedDC can be formulated as follows:
\begin{equation}
\min \limits_{\{P_k, A_k\}_{k=1}} \sum\limits_{k=1}^{K} \frac{n_k}{n} \mathcal{L}_{\mathcal{D}_k}(P_k, A_k),
\end{equation}
with $\mathcal{L}_{\mathcal{D}_k}(P_k, A_k)=\mathbb{E}_{(x, {y}) \in {D}_k} {\ell}_{ce}(f(P_k, A_k;x),y)$, where ${\ell}_{ce}$ denotes the cross-entropy loss function.

\section{Experiments}
In this section, we conduct comprehensive experiments to numerically evaluate pFedDC in scenarios involving complex data heterogeneity, including label and domain shifts.
First, in Sec.\ref{sec: experiment setup}, we provide a detailed introduction to our datasets, baselines, and experimental implementation details. 
Then, in Sec.\ref{sec: results}, we present our experimental results along with an ablation study for cross-attention modules.

\subsection{Experimental Setup}
\label{sec: experiment setup}

\textbf{Datasets.}
\label{Datasets}
We evaluate the performance of our method on nine public benchmark datasets characterized by three types of data heterogeneity: label shift, domain shift, and both shifts exist simultaneously.
Building on previous researchs~\cite{li2024global,cui2024harmonizing}, we utilize seven representative visual classification datasets: 
\textbf{Caltech101}~\cite{fei2007learning}, \textbf{Flowers102}~\cite{nilsback2008automated}, \textbf{OxfordPets}~\cite{parkhi2012cats}, 
\textbf{DTD}~\cite{cimpoi2014describing}, 
\textbf{CUB}~\cite{wah2011caltech},    \textbf{UCF101}~\cite{soomro2012ucf101}, and \textbf{EuroSAT}~\cite{helber2019eurosat} to simulate label shift with Dirichlet distribution.
Specifically, clients receive samples for each class based on a Dirichlet distribution in the case of label shift.
Here, the parameter $\beta$ controls the degree of label skew, with lower values indicating severe label skew.
For domain shift, we evaluate our framework on the following two datasets, each containing multiple domains: \textbf{DomainNet}~\cite{peng2019moment} with 6 domains including Clipart (\textbf{C}), Infograph (\textbf{I}), Painting (\textbf{P}), Quickdraw (\textbf{Q}), Real (\textbf{R}), Sketch (\textbf{S}) and \textbf{OfficeCaltech10}~\cite{gong2012geodesic} with 4 domains including Amazon (\textbf{A}), Caltech (\textbf{C}), DSLR (\textbf{D}), Webcam (\textbf{W}), where data from one of these distinct domains is assigned to one client.
For label and domain shifts, we use a Dirichlet distribution with $(\beta = 0.1)$ to distribute the samples of each domain across 5 clients in both the DomainNet and Office-Caltech10 datasets.

\begin{table}[t]
\setlength{\tabcolsep}{2pt}
\centering
\caption{The results (\%) of pFedDC and the benchmark methods on various datasets with Dirichlet-based ($\beta=0.1$) label shift. 
\textbf{Bold} and \underline{underline} represent the best and second-best results, respectively.}
\label{tab: label shift 0.1}
\resizebox{.95\textwidth}{!}{ 
\begin{tabular}{lcccccccc}
\toprule 
\textbf{Datasets} & {Caltech101} & {Flowers102} & {OxfordPets} & {DTD} & {CUB} & {UCF101} & {EuroSAT} & \textbf{Avg.}  \\ \midrule
Zero shot CLIP~\cite{radford2021learning} & 86.95 & 65.72 & 85.02 & 44.22 & 51.28 & 61.00 & 53.79& 60.57 \\ 	
PromptFL~\cite{guo2023promptfl}  & 90.51 & 73.62 & 91.55 & 52.10 & 68.63 & 84.03 & 90.83 & 78.75 \\
Promptprox~\cite{li2020federated} & 89.96 & 73.20 & 91.06 & 52.83 & 68.77 & 84.98 & 90.17 & 78.71 \\	
pFedPrompt~\cite{guo2023pfedprompt} & 96.02 & 86.75 & 92.30 & 80.59 & 83.04 & 86.22 & 92.49 & 88.20 \\		
FedOTP~\cite{li2024global} & \underline{96.44} & \underline{97.28} & \underline{98.94} & \textbf{92.77} & \underline{85.09} & \underline{87.56} & \underline{93.33} & \underline{93.06} \\\cmidrule{1-9} \rowcolor[gray]{0.9}
\textbf{pFedDC} & \textbf{99.27} & \textbf{99.45} & \textbf{99.24} & \underline{91.42} & \textbf{87.19} & \textbf{89.63} & \textbf{94.79} & \textbf{94.43} \\ \bottomrule
\end{tabular}
}
\end{table}

\begin{table}[t]
\setlength{\tabcolsep}{2pt}
\centering
\caption{The results (\%) of pFedDC and the benchmark methods on various datasets with Dirichlet-based ($\beta=0.05$) label shift. 
\textbf{Bold} and \underline{underline} represent the best and second-best results, respectively.}
\label{tab: label shift 0.05}
\resizebox{.95\textwidth}{!}{ 
\begin{tabular}{lcccccccc}
\toprule 
\textbf{Datasets} & {Caltech101} & {Flowers102} & {OxfordPets} & {DTD} & {CUB} & {UCF101} & {EuroSAT} & \textbf{Avg.}  \\ \midrule
Zero shot CLIP~\cite{radford2021learning} & 86.95 & 65.72 & 85.02 & 44.22 & 51.28 & 61.00 & 53.79& 60.57 \\ 	
PromptFL~\cite{guo2023promptfl}  & 88.72 & 70.02 & 89.26 & 50.46 & 68.41 & 82.72 & 88.76 & 86.90 \\
Promptprox~\cite{li2020federated} & 88.10 & 70.41 & 88.79 & 51.11 & 68.60 & 83.70 & 88.02 & 86.96 \\	
pFedPrompt~\cite{guo2023pfedprompt} & 95.88 & 82.90 & 91.27 & 77.92 & 82.86 & 85.49 & 91.60 & 86.84 \\		
FedOTP~\cite{li2024global} & \underline{97.12} & \underline{96.50} & \underline{98.45} & \underline{91.86} & \underline{85.11} & \underline{87.74} & \underline{92.69} & \underline{92.78} \\\cmidrule{1-9}  \rowcolor[gray]{0.9}
\textbf{pFedDC} & \textbf{99.31} & \textbf{99.28} & \textbf{99.30} & \textbf{91.95} & \textbf{87.03} & \textbf{90.51} & \textbf{94.85} & \textbf{94.60} \\ \bottomrule
\end{tabular}
}
\end{table}

\textbf{Baselines.}
In our experiments, we compare pFedDC with five baselines: (1) Zero-shot CLIP~\cite{radford2021learning} with hand-crafted text prompt templates “a photo of a [CLASS]”; (2) PromptFL~\cite{guo2023promptfl} using unified prompts learned across clients using FedAvg~\cite{mcmahan2017communication}; (3) Promptprox~\cite{li2020federated} integrate PromptFL~\cite{guo2023promptfl} with Fedprox~\cite{li2020federated};
(4) pFedPrompt~\cite{guo2023pfedprompt} equips each client with a non-parametric personalized attention module that synthesizes spatially adaptive visual prompts.
(5) FedOTP~\cite{li2024global} employs unbalanced optimal transport to align local visual features with textual features.
PromptFL~\cite{guo2023promptfl} and Promptprox~\cite{li2020federated} simultaneously optimize both the text and vision prompts, whereas FedOTP~\cite{li2024global} optimizes only the text prompts.

\textbf{Implementation details.}
We adopt the widely recognized vision-language model CLIP ViT-B/16 as our base model.
For label shift, we set the number of clients K to 10; for domain shift, we assign 1 client per domain; and for label and domain shift, we assign 5 clients per domain.
We conduct 20 communication rounds for all experimental datasets, regenerating pseudo labels using the updated local model every 5 communication rounds.
We optimize the prompts and cross-attention modules using mini-batch Stochastic Gradient Descent (SGD) with a learning rate of 0.01. 
For each experimental setting, we conduct three trials and report the mean accuracy.
All experiments are conducted with Pytorch~\cite{paszke2019pytorch} on NVIDIA 3090 GPUs.

\begin{table}[t]
\setlength{\tabcolsep}{2pt}
\centering
\caption{The results (\%) of pFedDC and the benchmark methods on Office-Caltech10 and DomainNet datasets with domain shift. \textbf{Bold} and \underline{underline} represent the best and second-best results, respectively.}
\label{tab: feature shift}
\resizebox{.95\textwidth}{!}{ 
\begin{tabular}{lcccccccccccc}
\toprule 
\textbf{Datasets} & \multicolumn{5}{c}{{Office-Caltech10}} & \multicolumn{7}{c}{{DomainNet}}  \\ \cmidrule(lr){2-6}  \cmidrule(lr){7-13}
\textbf{Domains} & {\textbf{A}} & {\textbf{C}} & {\textbf{D}} & {\textbf{W}} & {\textbf{Avg.}} & {\textbf{C}} & {\textbf{I}} & {\textbf{P}} & {\textbf{Q}} & {\textbf{R}} & {\textbf{S}} & {\textbf{Avg.}} \\ \midrule
Zero shot CLIP~\cite{radford2021learning}  & 9.05 & 7.23 & 15.45 & \underline{7.92} & 9.91  &  11.82 &  8.80 & 14.29 & 5.44 & 10.17 & 14.34 & 10.81 \\ 	
PromptFL~\cite{guo2023promptfl}  & 95.57 & 95.38 & 97.48 & \textbf{100.00} & 97.11  &  98.62 & 79.33 & 97.60 & 85.79 & 97.76 & 97.71 & 92.82 \\
Promptprox~\cite{li2020federated} & 96.14 & 95.84 & \underline{98.62} & \textbf{100.00} & 97.65 &   98.58 & 78.39 & 98.00 & 81.92 & 98.06 & 97.14 & 92.02 \\	
pFedPrompt~\cite{guo2023pfedprompt} & \underline{97.01} & 96.31 & \textbf{100.00} & \textbf{100.00} & \underline{98.33} & \underline{98.63} & \underline{81.25} & 98.42 & \underline{88.57} & \underline{98.11} & \underline{98.30} & \underline{93.88} \\
FedOTP~\cite{li2024global} & {96.42} & \underline{96.38} & \textbf{100.00} & \textbf{100.00} & {98.20} & {97.95} & {78.57} & \underline{98.44} & {84.43} & {97.60} & {97.87} & {92.48} \\\cmidrule{1-13}  \rowcolor[gray]{0.9}
\textbf{pFedDC} & \textbf{97.46} & \textbf{97.89} & \textbf{100.00} & \textbf{100.00} & \textbf{98.84}   & \textbf{98.70} & \textbf{84.49} & \textbf{99.10} & \textbf{93.08} & \textbf{98.25} & \textbf{98.76} & \textbf{95.40} \\ \bottomrule
\end{tabular}
}
\end{table}

\begin{table}[t]
\setlength{\tabcolsep}{2pt}
\centering
\caption{The results (\%) of pFedDC and the benchmark methods on Office-Caltech10 and DomainNet datasets with domain $\&$ label shifts. \textbf{Bold} and \underline{underline} represent the best and second-best results, respectively.}
\label{tab: feature and label shift}
\resizebox{.95\textwidth}{!}{ 
\begin{tabular}{lcccccccccccc}
\toprule 
\textbf{Datasets} & \multicolumn{5}{c}{{Office-Caltech10}} & \multicolumn{7}{c}{{DomainNet}}  \\ \cmidrule(lr){2-6}  \cmidrule(lr){7-13}
\textbf{Domains} & {\textbf{A}} & {\textbf{C}} & {\textbf{D}} & {\textbf{W}} & {\textbf{Avg.}} & {\textbf{C}} & {\textbf{I}} & {\textbf{P}} & {\textbf{Q}} & {\textbf{R}} & {\textbf{S}} & {\textbf{Avg.}} \\ \midrule
Zero shot CLIP~\cite{radford2021learning}  & 9.05 & 7.23 & 15.45 & 7.92 & 9.91 & 11.82 & 8.80 & 14.29 & 5.44 &  10.17 & 14.34 & 10.81 \\ 	
PromptFL~\cite{guo2023promptfl}  & 15.01 & 14.33 & 23.16 & 19.29 & 17.95 & 9.98 & 17.06 & 8.07 & 13.31 & 19.06 & 10.66 & 13.02 \\
Promptprox~\cite{li2020federated} & 16.97 & 17.13 & 24.74 & 18.72 & 19.39 & 10.62 & 16.47 & 7.84 & 12.86 & 18.75 & 17.32 & 13.98 \\	
pFedPrompt~\cite{guo2023pfedprompt} & 23.88 & 28.98 & 36.81 & 30.64 & 30.07 & 37.58 & 50.19 & 35.32 & 33.19 & 41.27 & 39.75 & 39.55 \\	
FedOTP~\cite{li2024global} & \underline{25.54} & \underline{31.26} & \underline{41.95} & \underline{37.76} & \underline{34.13} & \underline{45.22} & \underline{55.62} & \textbf{47.06} & \underline{36.46} & \underline{48.64} & \underline{45.95} & \underline{46.49} \\\cmidrule{1-13} \rowcolor[gray]{0.9}
\textbf{pFedDC} & \textbf{27.36} & \textbf{35.34} & \textbf{46.33} & \textbf{40.29} & \textbf{37.33} & \textbf{46.31} & \textbf{56.89} & \underline{46.42} & \textbf{37.17} & \textbf{50.08} & \textbf{49.93} & \textbf{47.80} \\ \bottomrule
\end{tabular}
}
\end{table}

\subsection{Experimental Results}
\label{sec: results}

\textbf{Model evaluation on label shift.}
Table~\ref{tab: label shift 0.1} and Table~\ref{tab: label shift 0.05} show the performance results of various methods under Dirichlet-based label shift.
Except for the DTD~\cite{cimpoi2014describing} dataset under Dirichlet-based label shift with $\beta = 0.1$, our method performs better than the baseline on other datasets.
This demonstrates that our cross-attention module effectively fuses global and local prompts.
Notably, the performance of the personalized methods FedOTP~\cite{li2024global} and the proposed pFedDC significantly exceeds that of the traditional FL method, highlighting the substantial advantage of personalized federated learning approaches in the case of label shift.

\textbf{Model evaluation on domain shift.}
Table~\ref{tab: feature shift} presents the performance results of various methods on Office-Caltech10 and DomainNet datasets under domain shift conditions. 
Although FedOTP~\cite{li2024global} achieves better accuracy under the label-shift setting, it still underperforms compared to pFedPrompt~\cite{guo2023pfedprompt}, which utilizes personalized local visual embeddings to assist with prediction.
This indicates a significant gap between the visual embeddings generated by the frozen image encoder and the personalized text embeddings, highlighting the necessity of personalized processing of visual information.
However, our method significantly outperforms state-of-the-art algorithms on both datasets, demonstrating the effectiveness of our dual prompts tuning.
In particular, on the DomainNet~\cite{peng2019moment} dataset, our method improves upon pFedPrompt~\cite{guo2023pfedprompt} by approximately $2.5\%$.

\textbf{Model evaluation on label and domain shifts.}
In this experiment, we evaluate the performance of pFedDC in more challenging scenarios involving simultaneous label and domain shifts.
Specifically, we partition the data within each domain into five clients using a Dirichlet distribution with a concentration parameter $\beta = 0.1$, introducing both inter-client label imbalance and domain-level feature variation.
The evaluation is conducted on two benchmark datasets: DomainNet~\cite{peng2019moment} and OfficeCaltech10~\cite{gong2012geodesic}, with results summarized in Table~\ref{tab: feature and label shift}.
Conventional federated learning methods suffer from substantial accuracy degradation due to their limited capacity to adapt to both semantic and distributional divergences across clients.
In contrast, pFedDC demonstrates strong resilience, consistently outperforming baseline methods.
Notably, our approach achieves an average accuracy improvement of $1.3\%$ across all domains, underscoring its effectiveness in complex federated learning environments.

\begin{figure}[t]
  \centering
  \includegraphics[width=\textwidth]{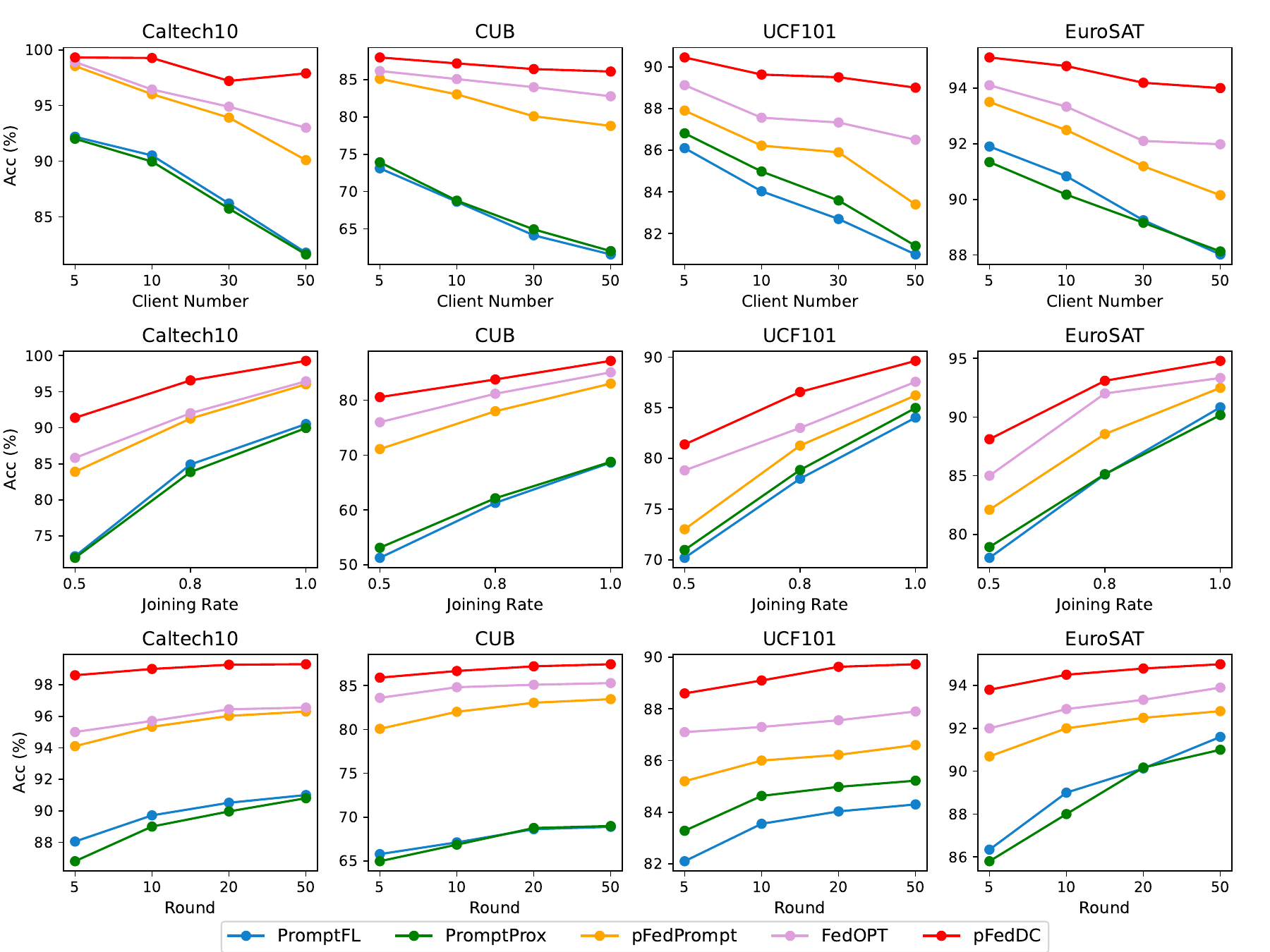}
  \caption{Experimental results under label skew settings with varying numbers of clients, client participation rates, and communication rounds.}
  \label{fig: client}
\end{figure}

\subsection{Analysis}
\textbf{Results under different client numbers.}
We evaluate the performance of the proposed pFedDC method in comparison with baseline approaches under varying numbers of participating clients.
Unless otherwise specified, all experiments are conducted using Dirichlet-based data partitioning with a concentration parameter of $\beta = 0.1$ to simulate non-IID data distributions.
Specifically, we partition the Caltech10, CUB, UCF101, and EuroSAT datasets among 5, 10, 30, and 50 clients, respectively.
The resulting classification accuracies are reported in the first row of Fig.~\ref{fig: client}.
As shown in the results, baseline methods exhibit a notable decline in performance as the number of clients increases, highlighting their sensitivity to data fragmentation and heightened heterogeneity.
In contrast, the proposed pFedDC consistently maintains high accuracy across all settings, demonstrating strong resilience to increasing client diversity.
This performance stability underscores the robustness of our dual-prompt learning strategy and cross-attention design, which effectively balances global knowledge sharing with local personalization. 
These findings confirm the scalability and adaptability of pFedDC in realistic federated learning scenarios involving large and diverse client populations.

\textbf{Results under different client participation rates.}
To investigate the robustness of the proposed method under varying client participation levels, we evaluate performance across participation rates of $\{0.5, 0.8, 1.0\}$.
As illustrated in the second row of Fig.~\ref{fig: client}, pFedDC consistently outperforms all competing methods across different rates.
A general decline in performance is observed for all approaches as the participation rate decreases, which is expected since reduced participation amplifies the divergence between the models of selected clients and the global model, leading to slower and less stable convergence.
Despite this challenge, our method exhibits significantly better stability and accuracy, demonstrating its resilience to partial client availability—a common scenario in practical federated deployments.

\textbf{Results under different communication rounds.}
In real-world federated learning scenarios, the number of available communication rounds is often constrained due to limited bandwidth, energy consumption, or time restrictions on client devices.
To evaluate the communication efficiency of the proposed pFedDC framework, we conduct experiments with varying communication rounds, specifically $T \in \{5, 10, 20, 50\}$.
As presented in the third row of Fig.~\ref{fig: client}, pFedDC consistently achieves superior performance compared to baseline methods across all communication budgets.
Notably, even under severely restricted conditions (e.g., $T=5$), our approach demonstrates a clear performance advantage, indicating its ability to converge more effectively and efficiently with fewer communication rounds.
This robustness is attributed to the lightweight nature of prompt tuning and the design of our dual-prompt learning with cross-attention, which enables faster knowledge alignment between local and global models.
These results underscore the practical applicability of pFedDC in real-world deployments, where communication efficiency is critical.

\begin{table}[t]
\setlength{\tabcolsep}{2pt}
\centering
\caption{The results (\%) of pFedDC and the benchmark methods on various datasets under Dirichlet-based ($\beta=0.1$) label shift with RN50 as the image encoder backbone. 
\textbf{Bold} and \underline{underline} represent the best and second-best results, respectively.}
\label{tab: label shift rn50}
\resizebox{.95\textwidth}{!}{ 
\begin{tabular}{lcccccccc}
\toprule 
\textbf{Datasets} & {Caltech101} & {Flowers102} & {OxfordPets} & {DTD} & {CUB} & {UCF101} & {EuroSAT} & \textbf{Avg.}  \\ \midrule
Zero shot CLIP~\cite{radford2021learning} & 81.10 & 64.91 & 84.24 & 40.79 & 48.75 & 60.29 &  44.73 & 60.57 \\ 	
PromptFL~\cite{guo2023promptfl}  & 82.23 & 72.74 & 90.52 & 50.60 & 65.48 & 80.73 & 88.23 & 78.75 \\
Promptprox~\cite{li2020federated} & 81.84 & 71.93 & 90.90 & 50.57 & 65.39 & 80.81 & 87.55 & 78.71 \\	
pFedPrompt~\cite{guo2023pfedprompt} & 87.50 & 86.25 & 91.79 & 77.39 & 80.22 & 81.29 & 89.93 & 88.20 \\		
FedOTP~\cite{li2024global} & \underline{88.87} & \underline{96.10} & \underline{98.01} & \underline{90.12} & \underline{83.17} & \underline{82.80} & \underline{91.42} & \underline{93.06} \\\cmidrule{1-9}  \rowcolor[gray]{0.9}
\textbf{pFedDC} & \textbf{90.64} & \textbf{98.36} & \textbf{98.72} & \textbf{90.26} & \textbf{85.73} & \textbf{85.29} & \textbf{92.04} & \textbf{94.43} \\ \bottomrule
\end{tabular}
}
\end{table}

\begin{table}[t]
\centering
\caption{\textbf{Ablation study}. Accuracies (\%) under Dirichlet-based label shift. T.Att. and I.Att. denote the text cross-attention and the image cross-attention.}
\label{tab: ablation}
\resizebox{0.95\textwidth}{!}{ 
\begin{tabular}{cccccccccc}
\toprule
T.Att. & I.Att.  & {Caltech101} & {Flowers102} & {OxfordPets} & {DTD} & {CUB} & {UCF101} & {EuroSAT} & \textbf{Avg.} \\ \midrule
- & - & 90.51 & 73.62 & 91.55 & 52.10 & 68.63 & 84.03 & 90.83 & 78.75 \\
- & \cmark & 98.25 & 98.69 & 99.04 & 91.05 & 87.01 & 89.49 & 94.55 & 94.01  \\
\cmark & - &  98.06 & 98.28 & 98.89 & 89.83 & 86.64 & 88.90 & 94.02 & 93.51 \\
\cmark & \cmark & 99.27 & 99.45 & 99.24 & 91.42  & 87.19 & 89.63 & 94.79 & 94.43 \\ 
\bottomrule
\end{tabular}
}
\end{table}

\textbf{Results under different image encoder backbones.}
To assess the generalizability and practical applicability of pFedDC, we conduct additional experiments using different image encoder architectures.
In particular, we evaluate model performance when employing the ResNet-50 (RN50) backbone, which is widely adopted in resource-constrained environments due to its moderate computational cost.
The comparison results under this setting are reported in Table~\ref{tab: label shift rn50}.
Across all evaluated datasets, pFedDC consistently outperforms competing methods, even when deployed with a relatively lightweight encoder.
This performance advantage highlights the adaptability of our dual-prompt learning strategy, which effectively enhances feature representation and learning stability despite the limited expressive capacity of smaller backbones.

\textbf{Ablation study.}
To investigate the contribution of the proposed cross-attention modules, we conduct an ablation study focusing on their impact during local training.
Specifically, we evaluate the performance under three settings: using only the text cross-attention module, using only the vision cross-attention module, and disabling both modules.
All experiments are performed under Dirichlet-based label shift conditions to simulate realistic non-IID scenarios.
The results, summarized in Table~\ref{tab: ablation}, indicate that introducing cross-attention in either modality leads to a notable improvement in accuracy compared to the setting without any cross-attention. 
Furthermore, applying cross-attention in both text and vision modalities simultaneously yields the highest performance, confirming the complementary benefits of modality-specific fusion.
These findings validate the effectiveness of our cross-attention design in integrating global and local prompts, thereby enhancing model personalization and generalization in federated learning environments.

\section{Conclusion}
This work addresses the challenge of complex data heterogeneity in federated learning, where domain shift and label shift occur simultaneously. 
We propose personalized federated learning with dual-prompt learning and cross fusion module (pFedDC), which incorporates both text and vision prompt learning along with personalized local cross-attention modules for each modality.
Our pFedDC enhances efficient collaborative prompt learning, enabling it to capture diverse categories and domain traits on a per-client basis while effectively learning valuable knowledge from other clients.
Extensive experiments on datasets with various types of heterogeneity demonstrate that our pFedDC outperforms state-of-the-art methods.
In the future, we will conduct a theoretical analysis of pFedDC, focusing on convergence, privacy, fairness, and other relevant considerations.
\bibliographystyle{IEEEtran}
\bibliography{mir-article.bbl}

\end{document}